\documentclass{article}

\usepackage{arxiv}
\usepackage[utf8]{inputenc} 
\usepackage[T1]{fontenc}    
\usepackage{graphicx}      
\usepackage{amsfonts}
\usepackage{amssymb}
\usepackage{amsmath}
\usepackage{natbib}        
\usepackage{booktabs}
\usepackage{wrapfig}
\usepackage{nicefrac}       
\usepackage{microtype}      
\usepackage{lipsum}
\usepackage{float}
\usepackage{array}
\usepackage{hyperref}
\usepackage[nolist]{acronym}
\begin{document}

\title{A Quantile Regression Approach for Remaining Useful Life Estimation with State Space Models}



%

\author{
  Davide Frizzo \\
  \texttt{davide.frizzo.1@phd.unipd.it} \\
  Department of Information Engineering \\
  University of Padova \\
  Padova (Italy) \\
  \And
  Francesco Borsatti \\
  \texttt{francesco.borsatti.1@phd.unipd.it} \\
  Department of Information Engineering \\
  University of Padova \\
  Padova (Italy) \\
  \And
  Gian Antonio Susto \\
  Department of Information Engineering \\
  University of Padova \\
  Padova (Italy) \\
  \texttt{gianantonio.susto@unipd.it} \\
}


\maketitle

\begin{abstract}                


\ac{PdM} is pivotal in Industry 4.0 and 5.0, proactively enhancing efficiency through accurate equipment \ac{RUL} prediction,
thus optimizing maintenance scheduling and reducing unexpected failures and premature interventions.
This paper introduces a novel \ac{RUL} estimation approach leveraging \ac{SSM} for efficient long-term sequence modeling.
To handle model uncertainty, \ac{SQR} is integrated into the \ac{SSM}, enabling multiple quantile estimations.
The proposed method is benchmarked against traditional sequence modelling techniques (\ac{LSTM}, Transformer, Informer)
using the \ac{C-MAPSS} dataset. Results demonstrate superior accuracy and computational efficiency of \ac{SSM} models,
underscoring their potential for high-stakes industrial applications.

\end{abstract}



\section*{}
\begin{acronym}
  \acro{AI}{Artificial Intelligence}
  \acro{ML}{Machine Learning}
  \acro{MSE}{Mean Squared Error}
  \acro{RMSE}{Root Mean Squared Error}
  \acro{MAE}{Mean Absolute Error}
  \acro{RL}{Reinforcement Learning}
  \acro{DL}{Deep Learning}
  \acro{LSTM}{Long Short Term Memory}
  \acro{GRU}{Gated Recurrent Unit}
  \acro{CNN}{Convolutional Neural Networks}
  \acro{PdM}{Predictive Maintenance}
  \acro{C-MAPSS}{Commercial Modular Aero-Propulsion System Simulation}
  \acro{RUL}{Remaining Useful Life}
  \acro{SQR}{Simultaneous Quantile Regression}
  \acro{CDF}{Cumulative Distribution Function}
  \acro{RNN}{Recurrent Neural Networks}
  \acro{SSM}{State Space Models}
  \acro{HiPPO}{High-order Polynomial Projection Operators}
  \acro{FFT}{Fast Fourier Transform}
  \acro{SISO}{Single Input Single Output}
  \acro{MIMO}{Multiple Input Multiple Output}
\end{acronym}

\section{Introduction} \label{sec:introduction}

In the rapidly evolving landscape of Industry 4.0 and Industry 5.0
(\cite{industry_5.0}), \ac{PdM} has emerged as a crucial component of smart
manufactoring , offering a more efficient and cost-effective approach to
maintenance. In opposition to traditional maintenance strategies, such as Run
to Failure and Preventive Maintenance, which rely on reactive measures,
\ac{PdM} addresses the problem in a proactive way, by maximizing the useful
life of the equipment while minimizing its downtime and maintenance costs. This
objective is achieved through continuous monitoring of machine health and
prediction of its \ac{RUL}, defined as the elapsed time to failure, thereby
enabling maintenance operations to be scheduled only when needed.

The timing of maintenance interventions is of paramount importance as it is
strictly related to the trade off between two undesirable scenarios: performing
maintenance too late, potentially leading to \textit{unexpected breaks}, or too
early, resulting in \textit{unexploited lifetime}. These two aspects are
usually quantified by the costs associated to the maintenance operations and
the equipment downtime in order to formulate ad-hoc business metrics to
evaluate the performance of the \ac{RUL} estimation (\cite{pdm_industry})
method.

Recent advancements in the field of \ac{ML} and \ac{DL} have enabled the
development of more accurate, reliable and efficient data-driven methods for
\ac{PdM}. This work investigates the applicability of \ac{SSM} (\cite{s4}), a
class of methods originally employed in control theory and recently adapted to
the field of \ac{DL}, to the \ac{RUL} estimation task. Specifically, we
leverage the outstanding long-term sequence modelling performance of \ac{SSM}
to improve \ac{RUL} estimation.

Quantifying model uncertainty is another crucial aspect of \ac{RUL} estimation
as it provides a quantitative and qualitative framework for visualizing the
trade-off between \textit{unexpected breaks} and \textit{unexploited lifetime}.
This enables users to make informed decisions by prioritizing one aspect over
the other, thereby modulating the degree of risk associated with maintenance
operations. This is achieved utilizing \ac{SQR}, which modifies the model
architecture and the training process to enable the design of a model able to
estimate multiple quantiles of the target distribution, resulting in a
comprehensive representation of uncertainty.

The remainder of this paper is organized as follows: Section
\ref{sec:related_work} presents an overview of the field of \ac{PdM} and
\ac{RUL} estimation, Section \ref{sec:method} introduces the proposed method,
Section \ref{sec:experiments} describes the experimental setup and discusses
the results. Finally in Section \ref{sec:conclusion} conclusions are drawn and
future works are outlined.

\section{Related Work} \label{sec:related_work}

As discussed in Section \ref{sec:introduction}, \ac{PdM} recently emerged as
a leading approach to manufacturing health management thanks to its proactive
nature, offering a significant improvement over traditional maintenance
techniques such as \textbf{Reactive Maintenance} which involves perfoming
maintenance after equipment failure, and \textbf{Preventive Maintenance} where
fixed time intervals are employed to schedule maintenance operations. By
predicting the \ac{RUL} of the equipment, \ac{PdM} aims to minimize machine
downtime and maximize equipment lifespan. This work focuses on this crucial
task, proposing a novel \ac{RUL} estimation model based on \ac{SSM}.

The field of \ac{RUL} estimation has been widely explored in the literature,
with a variety of approaches that can be broadly categorized into
\textbf{model-based} and \textbf{data-driven}. Model-based approaches heavily
rely on a pre-existing physical model of the system ,which can be complex to
design requiring in-depth knowledge of the its dynamics. In contrast,
data-driven methodologies laverage historical data to estimate the \ac{RUL} of
the machine without requiring domain expertise, making these class of
approaches highly flexible and easier to implement.The continuous development
in the fields of \ac{ML} and \ac{DL}, in particular sequence modelling, makes
data-driven approaches as the prominent approach for \ac{RUL} estimation.

\cite{RUL_CMAPSS} provides a comprehensive overview of the state of the art in
\ac{RUL} estimation focusing on the renowed \ac{C-MAPSS} benchmark dataset
which will also be employed in this study, as detailed in Section
\ref{subsec:data}. The top-perfoming architectures on \ac{C-MAPSS} are commonly
based on \ac{DL} approaches, combining the \ac{LSTM} and \ac{CNN} architectures
(\cite{LSTM_CNN}). These models leverage their exceptional capabilities in
modelling temporal and spatial dependencies in raw sensor measurement data from
the monitored equipments. More recently also attention based approaches
(\cite{dual_attention,Zhao_2022}) have been developed. These architectures offer
the advantage of highly parallelizable training but suffer from a quadratic
time complexity with respect to input sequence length. This is particularly
significant in \ac{RUL} estimation where the length of the run-to-failure
cycles can be highly variable.

The recently introduced \ac{SSM} models offer a promising alternative for the
\ac{RUL} estimation task thanks to their enhanced efficiency and capability of
modelling long term dependencies as proved by the results obtained on the Long
Range Arena (\cite{long_range_arena}) benchmark.

\ac{PdM} technologies are usually employed in high-stakes industrial
applications, making the quantification of model uncertainty an aspect of
paramount importance. Overly optimistic or pessimistic prediction can
dramatically increase the maintenance costs or the equipment downtime,
underscoring the importance of balancing the trade-off between
\textit{unexpected breaks} and \textit{unexploited lifetime}. Traditional
approaches to adress this challenge rely on on Bayesian \ac{ML} techniques
(\cite{bayesian_pdm}, \cite{bayesian_transformer}) which, however, are highly
dependent on the choice of the prior distribution and often employ
computationally expensive sampling algorithms which may not guarantee
convergence on high dimensional problems. In contrast, this work adopts a
quantile regression based approach to quantify model uncertainty which requires
minimal modification to the model architecture and training process and is
capable of estimating multiple quantiles of the conditional distribution of the
target variable given the input data.

\section{Proposed Methodology} \label{sec:method}
 

\subsection{State Space Models} \label{subsec:ssm}

\ac{SSM} are a class of models used in many research fields such as control
theory (\cite{kalman1960}). Recently, \ac{SSM} have been proposed as a more
efficient alternative (\cite{s4}) to attention based models
(\cite{transformer},\cite{informer}) for sequence learning. A significant limitation of
the Transformer model is in fact the computational complexity of its
self-attention mechanism, which scales quadratically with the sequence length,
rendering it inpractical for long sequences, such as those encountered in the
\ac{RUL} estimation task.

\ac{SSM} are based on the differential equation \ref{eq:state_space_eq} which
describes the transition from an input signal $u(t)$ to an output signal $y(t)$
through an internal state $x(t)$:

\begin{equation} \label{eq:state_space_eq}
\begin{cases}
    \dot{x}(t) = A x(t) + B u(t) \\
    y(t) = C x(t) + D u(t)
\end{cases}
\end{equation}

where $A$, $B$, $C$ and $D$ are the system matrices which, in this novel
implementation of \ac{SSM}, are learned from data using gradient-based
approaches. The adaptation of \ac{SSM} in Deep Learning is facilitated by the
\ac{HiPPO} framework which provides a smart initialization of matrix $A$ as the
sum of a low-rank and a normal term. The performances of \ac{SSM} models with
\ac{HiPPO} initialized matrices shows a significant improvement, from 60 \% to
90 \%, compared to randomly initialized matrices in the sequential MNIST task
(\cite{hippo}). Intuitively, the \ac{HiPPO} operator efficiently captures the
history of the input signal $u(t)$ in a finite state $x(t)$, representing the
coefficients of a polynomial reconstructing $u$ up to time $t$. The state
$x(t)$ is updated online as new time steps are considered, forming a
continuous-time memorization system which significantly improves the model
performances on long sequences.

To understand the efficiency of \ac{SSM} consider the discrete time version of
the state space equation:

\begin{equation} \label{eq:state_space_eq_discrete}
\begin{cases}
    x_{t+1} = A x_t + B u_t \\
    y_t = C x_t + D u_t
\end{cases}
\end{equation}

This equation has a strong resemblance to that of a \ac{RNN} (\cite{rnn}) with
the key difference being the absence of a non linear activation function, This
makes \ac{SSM} models a linear variant of \ac{RNN}. While \ac{RNN} models are
theoretically capable of capturing long-term dependencies due to their
unbounded context window, they suffer from the vanishing or exploding gradient
problem. Furthermore, their inherently sequential formulation renders the
learning process not parallelizable, leading to computationally inefficiency.

In contrast, the absence of a non linear term in \ac{SSM} enables parallel
training. Specifically, unfolding the state equation
\ref{eq:state_space_eq_discrete} in time yields a parallelizable convolutional
operation (\ref{eq:ssm_conv}) which generates the output signal $y_t$ from the
input signal $u_t$ without requiring computation of the hidden state $x_t$.
This leads to the convolutional view of \ac{SSM}, which underlies their
efficiency.

\begin{equation} \label{eq:ssm_conv}
  \begin{aligned}
    y_k &= \overline{CA^k}u_0 + \overline{CA^{k-1}}\overline{B}u_1 + \dots + \overline{CAB}u_{k-1} + \overline{B}u_k \\
    y &= \overline{K} * u
  \end{aligned}
\end{equation}


The convolution operator in \ref{eq:ssm_conv} is efficiently computed in the
frequency domain using the \ac{FFT} where convolution simplifies to
element-wise multiplication.

The \ac{SSM} models employed in this work are \texttt{S4}, \texttt{S4D},
\texttt{S5} (\cite{s4},\cite{s4d},\cite{s5}). The \texttt{S4} model is the simplest of the
three, leveraging the dual recurrent-convolutional nature of \ac{SSM} described
earlier. It is designed as a \ac{SISO} system, where each input feature is
processed independently by a separate \ac{SSM} block and a final mixing layer
to combine the outputs. The \texttt{S4D} model simplifies the \ac{HiPPO}
operator using diagonal matrices, which reduces computational complexity but
compromises theoretical viability (i.e. \ac{HiPPO} matrices are not
diagonalizable). The \texttt{S5} model is a simplified variant of \texttt{S4}
that replaces the \ac{SISO} architecture with a \ac{MIMO} architecture,
processing all input variables in parallel within a single \ac{SSM} block.

\subsection{Quantile Regression} \label{subsec:quantile_regression}

To address the distinct impacts of overestimations and underestimations of the
\ac{RUL}, particularly the tradeoff between \textit{unexpected breaks} and
\textit{unexploited lifetime}, a Quantile Regression approach
(\cite{quantile_reg_book}) was employed.

While Regression based algorithms, typically trained through conventional loss
functions such as \ac{MSE} or \ac{MAE}, are designed to predict the mean or
median of the target variable's distribution (i.e. the \ac{RUL} in this
context), a Quantile Regression approach focuses on estimating quantiles of the
conditional posterior distribution $p(y|x)$ where $y$ is the target variable
and $x$ is the input data.

Consequently, this method offers a more comprehensive picture of model
uncertainty, with an emphasis on aleatoric uncertainty. Furthermore, such
models allow for the selection of specific quantiles during inference depending
on the application requirements; for instance, quantiles higher than the median
may be preferred if overestimation of the target variable is deemed more
acceptable then underestimation, which, inversely, may be achieved by
predicting lower quantiles.

Various implementations of Quantile Regression are documented in the literature
(\cite{quantile_forest},\cite{quantile_nn},\cite{deep_quantile}). In this study, \ac{SQR}
(\cite{quantile_regression}), whose aim is to provide a simoultaneous estimation
of multiple quantiles, was employed.

The starting point to achieve such an objective is the \ac{CDF} of the target
varible $Y$, expressed as $F(y) = P(Y \leq y)$. Building on this the
\textbf{quantile distribution function} for a specific quantile level $\tau \in
[0,1]$ can be defined as:

\begin{equation} \label{eq:quantile_dist}
F^{-1}(\tau) = \text{inf} \ \{y | F(y) \geq \tau \}
\end{equation}

Given the input data $x \in \mathbb{R}^n$, the task involves estimating a
particular quantile $\tau$ of $Y$ given $x$, thereby constructing a model
$\hat{y} = \hat{f}_{\tau}(x)$ to approximate the \textbf{conditional quantile
distribution function} $y = F^{-1}(\tau|X=x)$.

It can be demonstrated that the optimal loss function for developing such a
model is the \textit{Pinball Loss}:

\begin{equation} \label{eq:pinball_loss}
\mathcal{L}_{\tau}(y,\hat{y}) =
\begin{cases}
        \tau(y - \hat{y}) \quad \text{if} \ y \geq \hat{y} \\
        (1-\tau)(\hat{y} - y) \quad \text{otherwise}
\end{cases}
\end{equation}

This cost function is defined for a single quantile level $\tau$, so if used to
train a regression model, it will yield an estimate for a single percentile. To
extend this to multiple quantiles, the \ac{SQR} approach assigns a randomly
sampled quantile $\tau_i$ level to each training sample $x_i$ and evaluates the
model performances on the new prediction $y_i=f(x_i, \tau_i)$ with
$\mathcal{L}_{\tau_i}(y_i, \hat{y}_i)$.

Formally, the training objective is:

\begin{equation} \label{eq:quantile_loss}
    \hat{f} \in \text{argmin}_{f} \frac{1}{n} \sum_{i=1}^{n} \mathbb{E}_ {\tau \sim U(0,1)} [\mathcal{L}_{\tau}(f(x_i,\tau),y_i)]
\end{equation}

where $U(0,1)$ denotes the uniform distribution under the interval $[0,1]$.

\subsection{Quantile Regression based Sequence Models for \ac {RUL} Estimation} \label{subsec:architecture}

By integrating concepts introduced in Sections \ref{subsec:ssm} and
\ref{subsec:quantile_regression}, we introduce a novel methodology, with the
aim of adapting \ac{SSM} based models to \ac{RUL} estimation exploiting
Quantile Regression to estimate the uncertainty of model predictions.

A conventional encoder-decoder architecture was employed for all the sequence
learning models compared in Section \ref{sec:experiments}. The architecture
comprises the following building blocks:

\begin{itemize}
  \item \textbf{Encoder}: The encoder projects the input sequence into an higher-dimensional latent space, which defines the input size
    of the sequence learning feature extraction backbone.
  \item \textbf{Feature Extraction Backbone}: This block is responsible for extracting the relevant features from the input sequence. Various
    sequence learning backbones were considered: \texttt{LSTM}, \texttt{Transformer}, \texttt{Informer} and \ac{SSM} models.
  \item \textbf{Decoder}: The decoder down-projects the extracted features into an output space, representing a signal
    containing \ac{RUL} predictions for each time step in the input sequence.
\end{itemize}

To account for the continuous-time nature of the input sensor measurement data,
a sliding window approach was employed to model inputs. Specifically, given an
input signal $\vec{x} = [x_0, x_1, \dots x_T] \in \mathbb{R}^T$, and the target
\ac{RUL} signal $\vec{RUL} = [RUL_0,RUL_1,\dots,RUL_T] = [T,T-1,\dots,0]$ with
a predefined window length $L$ the model input is constructed as follows: if
the signal's length is shorter than the window (i.e. $T < L$) it is zero-padded
to reach the desired length. Conversely, if the signal exceeds window length
(i.e. $T > L$) the input is constructed by sliding the window over the signal
with a stride of 1, thus resulting in $T-L+1$ sub sequences. Formally the input
signal is defined as:

$$
  \vec{x} =
  \begin{cases}
     [x_i,x_{i+1},\dots,x_{i+L-1}] \ \text{for} \ i \in [0,T-L+1] & \ \text{if} \ T > L \\
     [x_0,x_1,\dots,x_T,0,0,\dots,0] & \text{if} \ T < L
  \end{cases}
$$

The target \ac{RUL} signal corresponds to the selected window. If padding is
necessary for the input signal, the same procedure will be applied to the
target signal. A mask will then be employed within the loss function to ensure
that the model does not learn from these padded values.

$$
\vec{RUL} =
\begin{cases}
   [RUL_i,RUL_{i+1},\dots,RUL_{i+L-1}] \ \text{for} \ i \in [0,T-L+1] & \ \text{if} \ T > L \\
   [RUL_0,RUL_1,\dots,RUL_{T-1},0,0,\dots,0] & \text{if} \ T < L
\end{cases}
$$

Finally model's predictions across different windows are aggregated averaging
predictions over time samples shared across multiple windows, yielding the
final \ac{RUL} signal.


Such a comprehensive analysis of the input signal was implemented because in
\ac{RUL} estimation , real-time responses are not essential since commonly the
degradation process leading the machine towards failure is quite long. This
allows for the possibility of feeding the model with additional information on
the input signal to better capture the long term dependencies in the data.


Finally to account for the uncertainty inherent in model predictions, the
\ac{SQR} approach was employed to estimate various quantiles of the \ac{RUL}
distribution. As delineated in Section \ref{subsec:quantile_regression} it is
necessary to incorporate a quantile level $\tau$ within the model architecture
for the estimation of the corresponding quantile of the target variable.


In particular, the quantile percentage is included as a multiplicative factor
to the model's output to adjust the prediction according to the quantile level.

%


\section{Experimental Results} \label{sec:experiments}

In this section the experimental results will be outline. The code used for the
implementation of the proposed approach can be found at:
\url{https://github.com/FrizzoDavide/chronos-pdm}

\subsection{Data} \label{subsec:data}

The architectures delineated in Section \ref{subsec:architecture} are evaluated
on the \ac{C-MAPSS} (\cite{CMAPSS}) benchmark dataset. This dataset is commonly
used in the literature (\cite{RUL_CMAPSS}) because of the extensive volume of
labelled data, a critical factor for the \ac{RUL} estimation models. Labelled
data are typically challenging to obtain in real-world industrial applications
because it necessitates gathering data up to the failure event.


Consequently, \ac{C-MAPSS} is a synthetic dataset generated by simulating a
turbofan engine degradation process. In each simulation, also referred to as a
\textbf{run-to-failure cycle} or \textbf{life}, begins with an initial engine
health state randomly sampled and progresses through a degradation process
until failure occurs.

The dataset is partitioned into a training set - comprising sensor measurements
available until the end of life (i.e. $\ac{RUL} = 0$) - and a test set - where
data are available until some time prior failure, mimiking a real-world
scenario where the \ac{RUL} estimation is performed on the basis of the
available data up to the current time.

The ground truth \ac{RUL} signal is constructued by assuming a linear
degradation process, meaning that the \ac{RUL} decreases linearly by one unit
at each time step: $RUL=[T, T-1, T-2, \dots, 0]$, where $T$ is the length of a
run-to-failure cycle. This is a simplified assumption which is however widely
used in the literature (\cite{linear_degradation_rul}).

Finally, each dataset comprises 24 features, 3 being the operations conditions
and the remaining 21 being sensor measurements.

\begin{figure}[htbp]
    \centering
    \includegraphics[width=0.5\textwidth]{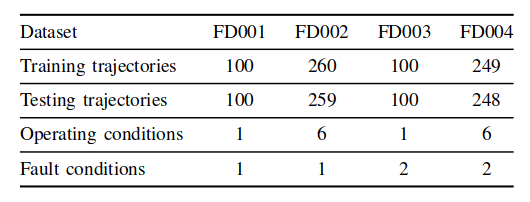}
    \caption{CMAPSS dataset subdivision}
    \label{fig:cmapss}
\end{figure}

The dataset is furtherly divided into four subset (i.e. \texttt{FD001},
\texttt{FD002}, \texttt{FD003}, \texttt{FD004}), as depicted in Figure
\ref{fig:cmapss}. These subsets vary by different numbers of faults and
operating conditions, thereby increasing the complexity of the \ac{RUL}
prediction task. Additional challenges include the heterogenity in the lengths
of run-to-failure cycles and the alterations in engine characteristics due to
manufacturing variations.

In this work we focused on the \texttt{FD001} and \texttt{FD002} datasets in
order to observe how the models perform when subjected to datasets
characterized by a different number of operating conditions.

\subsection{Results}

To evaluate the newly introduced \ac{SSM} models for \ac{RUL} using the
\texttt{C-MAPSS} dataset, we benchmarked them against other widely known
sequence modelling architectures. Within the \ac{RNN} family, from which
\ac{SSM} models are derived, we exclusively considered the \texttt{LSTM}
architecture (\cite{lstm}), as preliminary experiments on \ac{RNN} and \ac{GRU}
produced similar results. Additionaly, two attention based models,frequently
referenced in existing literature, were considered: the renowed
\texttt{Transformer} (\cite{transformer}) and the \texttt{Informer}
(\cite{informer}). The latter is an adaptation of the \texttt{Transformer}
architecture designed for long-term sequence modelling tasks such as time
series forecasting.


\newcolumntype{A}{>{\centering}p{0.09\textwidth}}
\newcolumntype{C}{>{\centering\arraybackslash}p{0.07\textwidth}}

\begin{table}[ht]
  \centering
  \begin{tabular}{|c|c|c|c|c|c|}
    \hline
    \textbf{Model} & \textbf{quantile 0.1} & \textbf{quantile 0.25} & \textbf{quantile 0.5} & \textbf{quantile 0.75} & \textbf{quantile 0.9} \\ \hline
    \texttt{S4} & 126.97 & 94.29 & 37.96 & \textbf{32.16} & \textit{54.38} \\ \hline
    \texttt{S5} & 128.07 & 93.91 & 48.60 & 43.49 & 71.68 \\ \hline
    \texttt{S4D} & 128.05 & 96.39 & 45.63 & 50.88 & 67.24 \\ \hline
    \texttt{LSTM} & 130.60 & 103.21 & 60.09 & \textit{35.92} & \textbf{41.26} \\ \hline
    \texttt{Transformer} & \textit{124.36} & \textit{87.44} & \textbf{31.46} & 41.07 & 75.09 \\ \hline
    \texttt{Informer} & \textbf{122.70} & \textbf{83.68} & \textit{32.41} & 54.77 & 79.83 \\ \hline
    \end{tabular}
  \vspace{0.7em}
  \caption{Average model performances on \ac{C-MAPSS} \texttt{FD001} across quantiles 0.1, 0.25, 0.5, 0.75 and 0.9}
  \label{tab:perf_FD001}
\end{table}

\begin{table}[ht]
  \centering
  \begin{tabular}{|c|c|c|c|c|c|}
    \hline
    Model         & quantile 0.1 & quantile 0.25 & quantile 0.5 & quantile 0.75 & quantile 0.9 \\ \hline
    \texttt{S4} & \textbf{127.02} & 88.10 & \textbf{35.60} & 62.50 & 99.66 \\ \hline
    \texttt{S5} & 134.55 & 103.17 & 53.56 & \textit{50.20} & \textit{94.07} \\ \hline
    \texttt{S4D} & 135.93 & 106.71 & 53.57 & 50.92 & \textbf{63.25} \\ \hline
    \texttt{LSTM} & 127.46 & \textit{86.45} & 36.11 & \textbf{36.11} & 96.45 \\ \hline
    \texttt{Transformer} & \textit{127.07} & \textbf{85.66} & \textit{35.59} & 62.49 & 99.65 \\ \hline
    \texttt{Informer} & 127.00 & 85.98 & 35.49 & 59.78 & 99.62 \\ \hline
    \end{tabular}
  \vspace{0.7em}
  \caption{Average model performances on \ac{C-MAPSS} \texttt{FD002} across quantiles 0.1, 0.25, 0.5, 0.75 and 0.9}
  \label{tab:perf_FD002}
\end{table}




Tables \ref{tab:perf_FD001} and \ref{tab:perf_FD002} report the test set
metrics for the \texttt{FD001} and \texttt{FD002} datasets, respectively. These
metrics are averaged over the various run-to-failure cycles within the test set
and over five experimental runs utilizing different random seeds to account for
the stochasticity of the training process.

In fact, the quantile levels are randomly sampled from a uniform distribution
over the interval $[0.1, 0.9]$. For each different evaluation quantile level
the two lowest error values are marked in bold and italic, respectively.

The employed evaluation criterion used is the \ac{RMSE} loss, commonly utilized
for \ac{RUL} estimation (\cite{RUL_CMAPSS}) due to its capacity to provide error
estimates on a scale comparable to that of the target variable, facilitating
result interpretation relative to other prevalent regression metrics such as
the \ac{MSE}.

A thorough examination of the test metrics revels how attention-based models
generally outperform others across most quantiles, particularly those below
0.5, closely followed by \ac{SSM} models. An analysis of the magniture of test
metrics across the different quantile outlines how extreme quantiles (i.e. 0.1
and 0.9) exhibit higher errors compared to intermediate ones due to a
significant underestimation or overestimation of the target variable.

The elevated errors observed in quantiles below the median likely stem from
models being trained on complete run-to-failure cycles ,where the \ac{RUL}
reaches 0, but evaluated on truncated lifes with fewer \ac{RUL} values near
zero. This discrepancy often results in frequent underestimation of the target
variable.





\begin{figure}[htbp]
    \centering
    \includegraphics[width=0.4\textwidth]{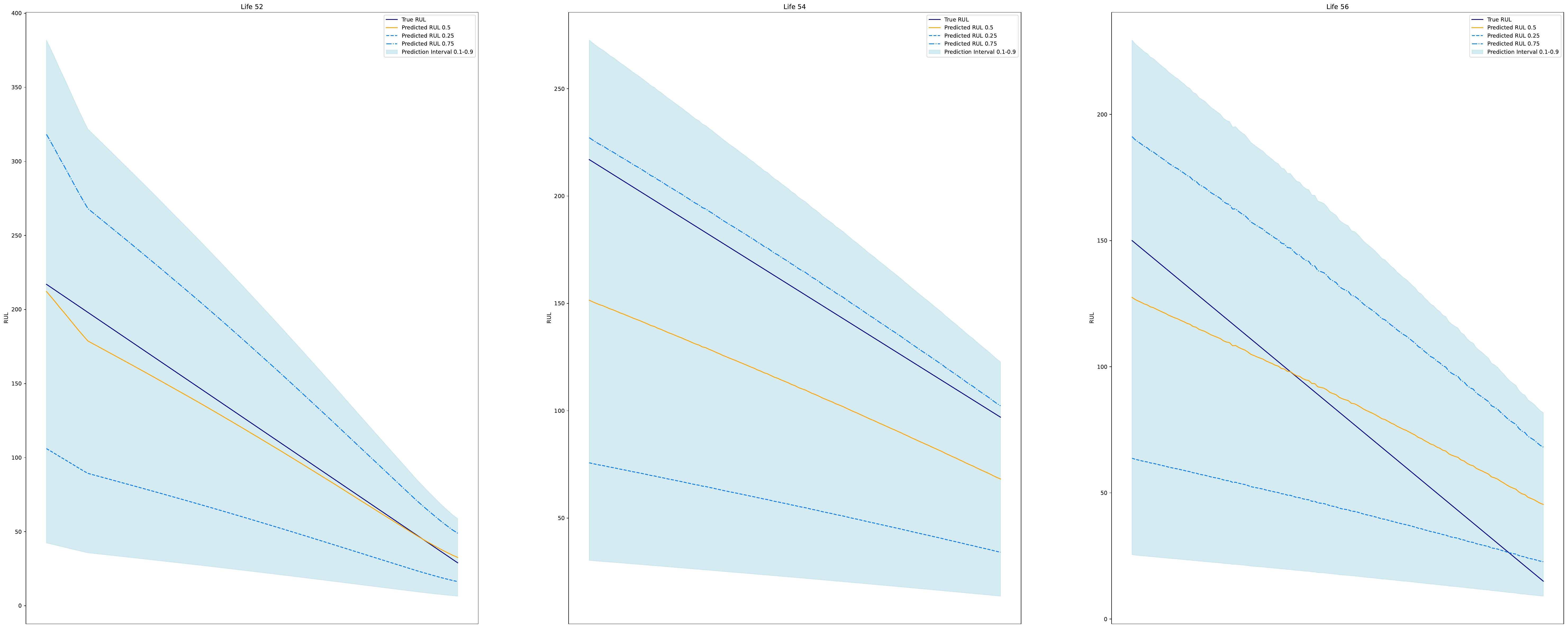}
    \caption{Prediction Interval plots on \texttt{FDOO1} produced by \texttt{S4} on lifes 52, 54, 56}
    \label{fig:RUL_S4}
\end{figure}



The \texttt{RUL} predictions obtained from the \texttt{FD001} dataset are
illustrated on Figure \ref{fig:RUL_S4}.

These plots depict model predictions across the evaluation quantiles:
$[0.1,0.25,0.5,0.75,0.9]$ juxtaposed with the actual \ac{RUL} signal. The
interval between predictions for the extreme quantiles 0.1 and 0.9 is shaded to
illustrate the model's prediction interval.

Analysing the graphs it is possible to discern the variability of the model's
predictions across different quantiles. The variability enables users to select
a desired confidence level in their predictions depending on specific
application needs. Quantiles exceeding 0.5 (the median of the target
distribution) can be laveraged to generate higher \ac{RUL} estimates, thereby
prioritizing machine usage over health with the risk of unexpected failures.
Conversely, adopting a more conservative approach involves selecting lower
quantiles to mitigate potential breakdown, even if the equipment is not fully
exploited.


\begin{figure}[htbp]
    \centering
    \includegraphics[width=0.4\textwidth]{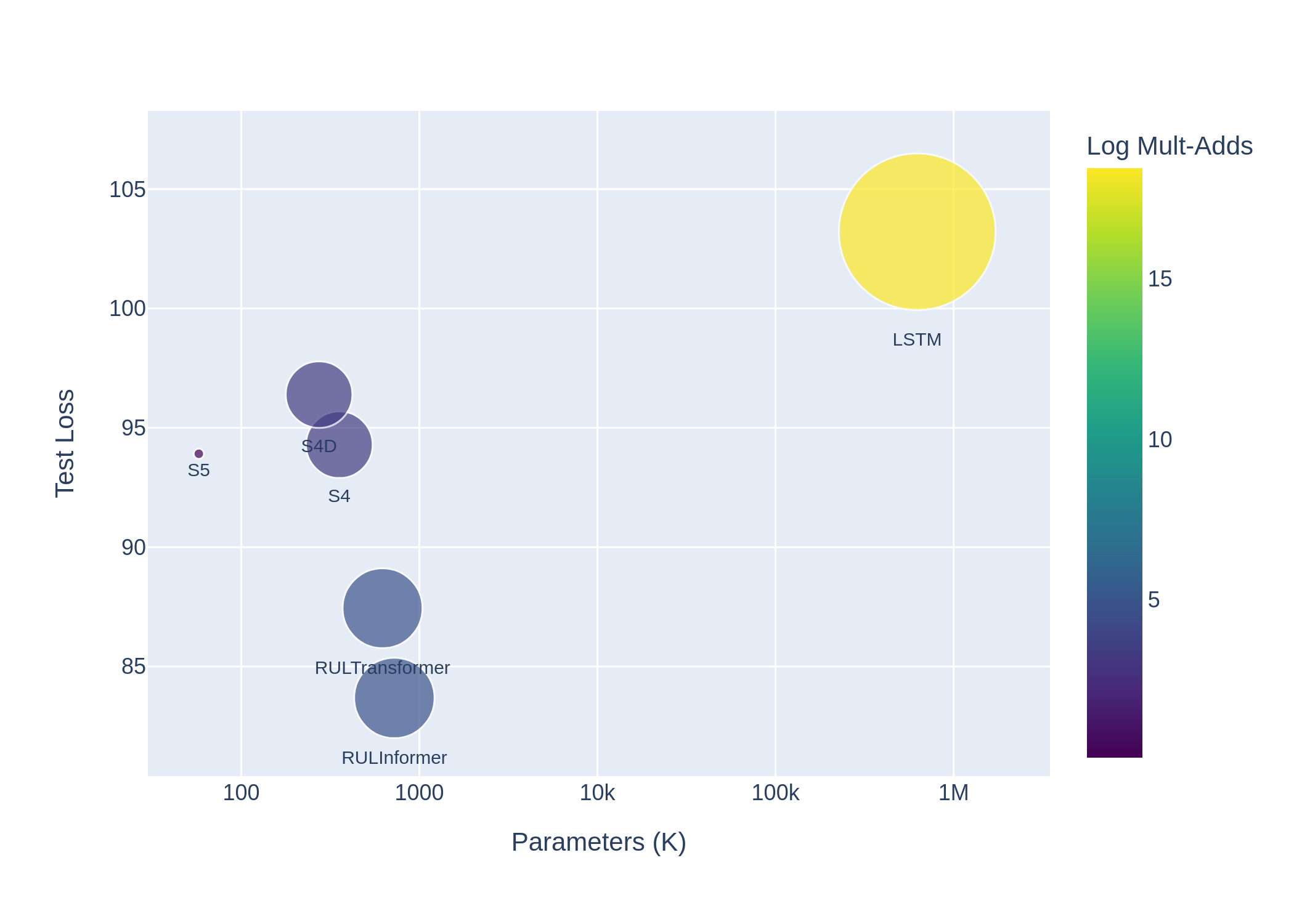}
    \caption{Blob plot for the comparison of model performances, model size and computational cost on \texttt{FD001} dataset}
    \label{fig:blob_FD001}
\end{figure}

A final comprehensive evaluation of the presented models considered three
crucial aspects for their applicability in industrial settings: model
performance, model size (i.e. number of parameters) and computational cost
(i.e. number of Mult-Adds operations). Figure \ref{fig:blob_FD001} depicts
various models as blobs, where the radius is proportional to the number of
Mult-Adds operations performed during a single forward pass. The horizontal and
vertical positions reflect model size and test set performance, respectively.

As previously outlined in Tables \ref{tab:perf_FD001},\ref{tab:perf_FD002},
\ac{SSM} and attention-based methods achieve comparable results in terms of
performance, thereby affirming their efficacy for \ac{RUL} estimation task.
However \ac{SSM} models showcase a significant reduction in both the number of
parameters and computational cost. This efficiency makes them better suited for
real-world applications, especially within environments where computational
resources are constrained, such as edge or cloud computing scenarios.


\section{Conclusion} \label{sec:conclusion}

This paper presented an innovative framework for \ac{RUL} estimation based on
\ac{SSM} enhanced with Quantile Regression techniques, effectively addressing
both the predictive accuracy and uncertainty quantification challenges inherent
in \ac{PdM}. Through rigorous experimental evaluation using the \ac{C-MAPSS}
dataset, the introduced approach consistently outperformed established sequence
modeling architectures, including \ac{LSTM} and performs on par with
Transformer and Informer models, across regression metrics. Furthemore the
\ac{SSM} architecture demonstrated superior computational efficiency, making it
a promising candidate for real-time applications in high-stakes industrial
environments.



The introduction of \ac{SQR} allows precise uncertainty estimation and provides
the flexibility to modulate the trade-off between \textit{unexpected breaks}
and \textit{unexploited lifetime} by selecting the desired quantile level for
the \ac{RUL} estimation based on the specific maintenance requirements of the
industrial system.

Future research directions includes validating the methodology on additional
\ac{PdM} benchmark datasets and on real-world industrial datasets to explore
the applicability of the proposed approach in practical scenarios.



\bibliographystyle{plain}

\begin{thebibliography}{10}

\bibitem{industry_5.0}
Aitzaz {Ahmed Murtaza}, Amina Saher, Muhammad {Hamza Zafar}, Syed {Kumayl Raza Moosavi}, Muhammad {Faisal Aftab}, and Filippo Sanfilippo.
\newblock Paradigm shift for predictive maintenance and condition monitoring from industry 4.0 to industry 5.0: A systematic review, challenges and case study.
\newblock {\em Results in Engineering}, 24:102935, 2024.

\bibitem{quantile_nn}
Alex Cannon.
\newblock Non-crossing nonlinear regression quantiles by monotone composite quantile regression neural network, with application to rainfall extremes.
\newblock {\em Environment and Climate Change Canada}, 12 2017.

\bibitem{rnn}
Jeffrey~L. Elman.
\newblock Finding structure in time.
\newblock {\em Cognitive Science}, 14(2):179--211, 1990.

\bibitem{deep_quantile}
Tobias Fissler, Michael Merz, and Mario~V. Wüthrich.
\newblock Deep quantile and deep composite triplet regression.
\newblock {\em Insurance: Mathematics and Economics}, 109:94–112, March 2023.

\bibitem{hippo}
Albert Gu, Tri Dao, Stefano Ermon, Atri Rudra, and Christopher Re.
\newblock Hippo: Recurrent memory with optimal polynomial projections, 2020.

\bibitem{s4}
Albert Gu, Karan Goel, and Christopher Ré.
\newblock Efficiently modeling long sequences with structured state spaces, 2022.

\bibitem{s4d}
Ankit Gupta, Albert Gu, and Jonathan Berant.
\newblock Diagonal state spaces are as effective as structured state spaces, 2022.

\bibitem{LSTM_CNN}
Yonghao He, Changjun Wen, and Wei Xu.
\newblock Residual life prediction of sa-cnn-bilstm aero-engine based on a multichannel hybrid network.
\newblock {\em Applied Sciences}, 15(2), 2025.

\bibitem{kalman1960}
Rudolph~Emil Kalman.
\newblock A new approach to linear filtering and prediction problems.
\newblock {\em Transactions of the ASME--Journal of Basic Engineering}, 82(Series D):35--45, 1960.

\bibitem{quantile_reg_book}
R.~Koenker.
\newblock {\em Quantile Regression}.
\newblock Cambridge University Press, 2005.

\bibitem{linear_degradation_rul}
Xiang Li, Wei Zhang, and Qian Ding.
\newblock Deep learning-based remaining useful life estimation of bearings using multi-scale feature extraction.
\newblock {\em Reliability Engineering and System Safety}, 182:208--218, 2019.

\bibitem{bayesian_pdm}
Luca~Della Libera, Jacopo Andreoli, Davide~Dalle Pezze, Mirco Ravanelli, and Gian~Antonio Susto.
\newblock Bayesian deep learning for remaining useful life estimation via stein variational gradient descent, 2024.

\bibitem{pdm_industry}
Luciano Lorenti, Davide~Dalle Pezze, Jacopo Andreoli, Chiara Masiero, Natalie Gentner, Yao Yang, and Gian~Antonio Susto.
\newblock Predictive maintenance in the industry: A comparative study on deep learning-based remaining useful life estimation.
\newblock In {\em 2023 IEEE 21st International Conference on Industrial Informatics (INDIN)}, pages 1--9, 2023.

\bibitem{quantile_forest}
Nicolai Meinshausen.
\newblock Quantile regression forests.
\newblock {\em Journal of Machine Learning Research}, 7(35):983--999, 2006.

\bibitem{lstm}
Hochreiter S. and Schmidhuber J.
\newblock Long short-term memory.
\newblock {\em Neural Computation}, 9(8):1735--1780, 1997.

\bibitem{CMAPSS}
Abhinav Saxena and Kai Goebel.
\newblock Turbofan engine degradation simulation data set, 2008.

\bibitem{s5}
Jimmy T.~H. Smith, Andrew Warrington, and Scott~W. Linderman.
\newblock Simplified state space layers for sequence modeling, 2023.

\bibitem{quantile_regression}
Natasa Tagasovska and David Lopez-Paz.
\newblock Single-model uncertainties for deep learning, 2019.

\bibitem{long_range_arena}
Yi~Tay, Mostafa Dehghani, Samira Abnar, Yikang Shen, Dara Bahri, Philip Pham, Jinfeng Rao, Liu Yang, Sebastian Ruder, and Donald Metzler.
\newblock Long range arena: A benchmark for efficient transformers, 2020.

\bibitem{transformer}
Ashish Vaswani, Noam Shazeer, Niki Parmar, Jakob Uszkoreit, Llion Jones, Aidan~N. Gomez, Lukasz Kaiser, and Illia Polosukhin.
\newblock Attention is all you need, 2017.

\bibitem{RUL_CMAPSS}
Simon Vollert and Andreas Theissler.
\newblock Challenges of machine learning-based rul prognosis: A review on nasa's c-mapss data set.
\newblock In {\em 2021 26th IEEE International Conference on Emerging Technologies and Factory Automation (ETFA )}, pages 1--8, 2021.

\bibitem{dual_attention}
Fan Wang, Aihua Liu, Chunyang Qu, Ruolan Xiong, and Lu~Chen.
\newblock A deep-learning method for remaining useful life prediction of power machinery via dual-attention mechanism.
\newblock {\em Sensors}, 25(2), 2025.

\bibitem{bayesian_transformer}
Feifan Xiang, Yiming Zhang, Shuyou Zhang, Zili Wang, Lemiao Qiu, and Joo-Ho Choi.
\newblock Bayesian gated-transformer model for risk-aware prediction of aero-engine remaining useful life.
\newblock {\em Expert Systems with Applications}, 238:121859, 2024.

\bibitem{Zhao_2022}
Zijian Zhao and Pengyuan Zou.
\newblock Attention-based dual-channel deep neural network for aero-engine rul prediction under time-varying operating conditions.
\newblock {\em Journal of Physics: Conference Series}, 2386(1):012027, dec 2022.

\bibitem{informer}
Haoyi Zhou, Shanghang Zhang, Jieqi Peng, Shuai Zhang, Jianxin Li, Hui Xiong, and Wancai Zhang.
\newblock Informer: Beyond efficient transformer for long sequence time-series forecasting.
\newblock {\em Proceedings of the AAAI Conference on Artificial Intelligence}, 35(12):11106--11115, May 2021.

\end{thebibliography}


\end{document}